\documentclass[lettersize,journal]{IEEEtran}
\usepackage{amsmath,amsfonts}
\usepackage{algorithmic}
\usepackage{algorithm}
\usepackage{array}
\usepackage[caption=false,font=normalsize,labelfont=sf,textfont=sf]{subfig}
\usepackage{textcomp}
\usepackage{stfloats}
\usepackage{url}
\usepackage{verbatim}
\usepackage{graphicx}
\usepackage{here}

\begin{document}

\title{OUXT Polaris: Autonomous Navigation System for the 2022 Maritime RobotX Challenge}
\author{
    Kenta Okamoto, Kyoto Institute of Tech., m2623106@edu.kit.ac.jp \\ \and
    Akihisa Nagata, Kansai Univ. , k065604@kansai-u.ac.jp \\ \and
    Kyoma Arai, Tokai Univ. , 2cemm007@mail.u-tokai.ac.jp \\ \and
    Yusei Nagao, Osaka Institute of Tech., m1m22r23@oit.ac.jp \\ \and
    Tatsuki Nishimura, Osaka Univ., hbvcg00@gmail.com \\ \and
    Kento Hirogaki, Gifu Univ., hkt8g2r6kin@gmail.com \\ \and
    Shunya Tanaka, syun111@gmail.com \\ \and
    Masato Kobayashi, Kobe Univ. , 171w951w@gsuite.kobe-u.ac.jp \\ \and
    Tatsuya Sanada, TXP Medical Co. Ltd., tatsuya.sh.tiny@gmail.com \\ \and
    Masaya Kataoka, TIER IV inc . , ms.kataoka@gmail.com
}

\markboth{Maritime RobotX Challenge 2022}%
{Shell \MakeLowercase{\textit{et al.}}: A Sample Article Using IEEEtran.cls for IEEE Journals}


\maketitle

\begin{abstract}
OUXT-Polaris has been developing an autonomous navigation system by participating in the 
Maritime RobotX Challenge 2014, 2016, and 2018. 
In this paper, we describe the improvement of the previous vessel system. 
We also indicate the advantage of the improved design.
Moreover, we describe the developing method under Covid-19 using simulation / miniture-size hardware and the 
feature components for the next RobotX Challenge.
\end{abstract}

\begin{IEEEkeywords}
Maritime systems, Robotics, Unmanned surface vehicle
\end{IEEEkeywords}

\section{Introduction}
\begin{figure}[H]
  \begin{center}
    \scalebox{0.4}{
      \includegraphics{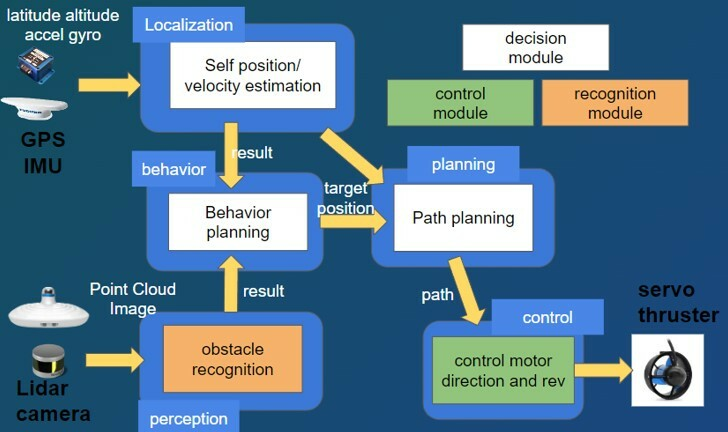}
    }
  \end{center}
  \caption{Achtecture of Navigation System}
  \label{fig:arch_nav}
\end{figure}

\begin{figure}[H]
  \begin{center}
      \scalebox{0.8}{
          \includegraphics{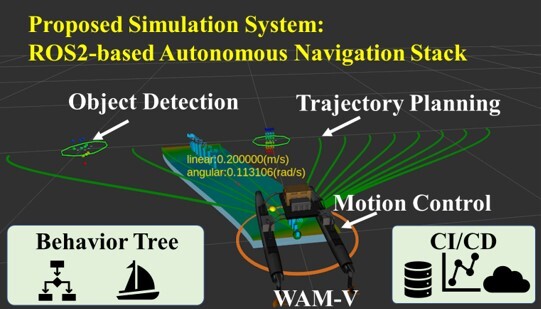}
      }
  \end{center}
  \caption{Proposed Software System}
  \label{fig:pss}
\end{figure}

First of all, we are motivated to develop a big field robot in a large area such as the ocean.
In recent years, the aging and shrinking population, as well as a shortage of workers,
has led to an increase in demand for the automation of cars, robots, and other equipment.
Among these, automated driving is being developed with particular emphasis.
Moving the autonomous vehicle or robot outside has a very severe problem.
They need to hedge unknown obstacles and go to the target position.
The environment such as weather, temperature, or underwater around robots causes sensor and hardware problems.
There are each challenging problems and They are also interesting for us, and there are different problems between land and ocean.
On the land, the navigation or the estimation of the self-position is solved by the point cloud map and the odmetry,
while on the ocean, the point cloud and the odmetry is not obtained enough. So the robots need to estimate the self-position using GPS and IMU sensors.
Moreover, on the land, the position of the target objects and obstacles is obtained from Lidar data. On the other hand in the sea,
waves disturbe to get the target positions. In that case, the robots have to fusion multiple data such as cameras and Lidars.
In this competition, we have a chance to develop a system to get over the wild environment 
for the robots on the ocean. Therefore, we are participating in the Maritime RobotX Challenge.
As shown in  Figs. \ref{fig:arch_nav}, and \ref{fig:pss},
we designed the architecture of the vessel navigation system. 
Our vessel navigation systems are composed of localization, perception, behavior, planning, and control.
Our localization, behavior, and planning methods are based on classical methods,
such as "Extended Kalman Filter", "Behavior Tree", "Cubic Hermite Spline" and "Velocity Control based on WAM-V Dynamics Model"
Our perception methods are based on learning methods, such as "YOLOX".
We beliefly describe the developed vessel navigation system as follows.

\section{Overview of vessel system}

\begin{enumerate}
  \item {\it Localization}: 
  The position and the velocity of WAM-V are estimated
  using the 6DoF Extended Kalman Filter \cite{robotx_ekf} from the data of the GNSS and IMU sensors.
    
  \item {\it Perception}: 
  The obstacles and task objects are recognized from lidar and camera data.
  We used YOLOX for object Detection of task object information such as buoys and docks.

  \item {\it Behavior}: 
  Using behavior tree can build WAM-V behaviors like a tree. 
  We use Groot \ref{fig:Groot}, GUI tools for designing behavior tree for smooth development.
  
  \item {\it Planning}: 
  Path planning generates obstacle-avoidable paths from sensors and WAM-V information in real-time.

  \item {\it Control}: 
  In the servo and thruster controllers,
  the servo motor direction and the thruster revolution
  to achieve the target velocity are calculated based on the vessel motion model.
\end{enumerate}

\section{Hardware Developments}

\subsection{Redesign of Azimuth Thruster}

We considered the following three issues when designing the propulsion mechanism.

First, we considered it important for the boat to be able to generate lateral propulsive force to complete the docking task.
When propulsion units are mounted on the two aft sections of the boat, each propulsion unit must have a degree of freedom in the yaw axis to generate thrust in any direction in the horizontal plane.
This propulsion system is generally called an azimuth thruster.

Next, the electric outboard motors that are commonly available have a circular cross-section for the mounting shaft, so it is necessary to find a way to fix the shaft tightly.

In addition, the propeller must avoid contact with the seafloor when the vessel needs to navigate in shallow water, such as when launching the boat on the course. Since it is dangerous for a person to enter shallow water and lift the thruster with a tool, a mechanism that can easily raise and lower the thruster was necessary.

\begin{figure}[H]
  \begin{center}
    \scalebox{0.2}{
      \includegraphics{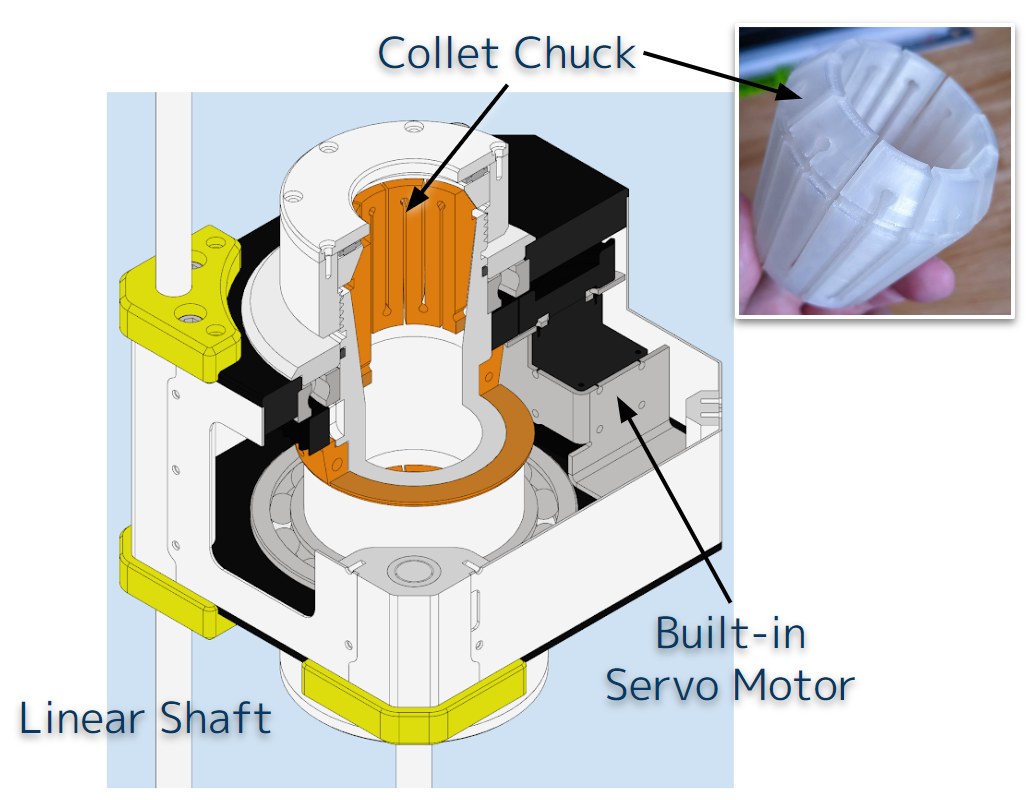}
    }
  \end{center}
  \caption{Design of Azimuth Thruster}
  \label{fig:azimuth_design}
\end{figure}

To meet these requirements, we designed the mechanism shown in Fig. \ref{fig:azimuth_design}.
The three functions of gripping, rotating, and elevating are integrated into a single unit.

The gripping function was realized using a PLA plastic collet manufactured by a 3D printer.
The collet is pushed axially by a screw into an aluminum hollow shaft with a wedge-shaped cross section to enable strong shaft gripping.

The azimuth mechanism is realized by transmitting the rotational force from the servo motor (XM540-W270, Dynamixel) to the hollow shaft by spur gears.
The hollow shaft is held at two points by angular bearings.

\subsection{Sensor Arrangement}

\begin{figure}[H]
  \begin{center}
    \scalebox{0.2}{
      \includegraphics{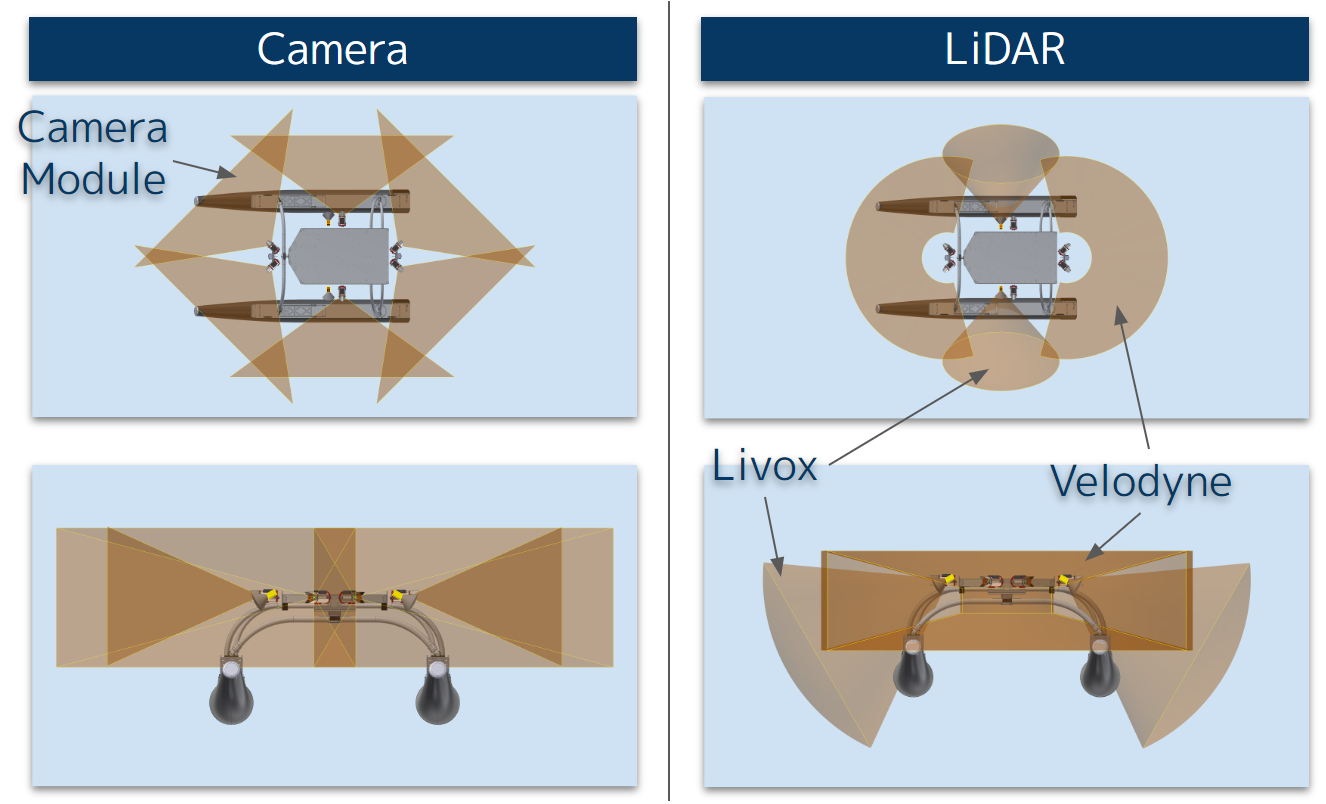}
    }
  \end{center}
  \caption{Sensor Arrangement}
  \label{fig:sensor_arrangement}
\end{figure}

LiDAR and a visible light camera are used for environmental awareness.
Their arrangement is shown in the Fig. \ref{fig:sensor_arrangement}.

In total, 6 cameras and 4 LiDARs are used.
Different LiDARs are used for the front and rear views and for the left and right views.
The VLP-16 from Velodyne Lidar is used for the front/rear view to provide a wide range of vision, mainly in the direction of boat travel, and the MID-70 from Livox is selected for the left/right view to see the docking bay near the hull in the docking task.
For the cameras, a module with an IMX-219 image sensor and a lens with a 120-degree diagonal field of view was used.
This allows the acquisition of point cloud information and visible light images covering 360 degrees around the boat.

\subsection{The Perception Array}

To perform point cloud fusion using a visible light camera and LiDAR, it is necessary to accurately calibrate the relative positions of the sensors.
Therefore, once the sensors are assembled on the hull, they cannot be easily removed for testing on land.

To solve this problem, we have developed a sensor unit that consists of a LiDAR and two cameras fixed to a rigid frame and can be mounted on or carried by various robots while maintaining the accuracy of the relative positioning between the sensors.
We call it a perception array.

\begin{figure}[H]
  \begin{center}
    \scalebox{0.1}{
      \includegraphics{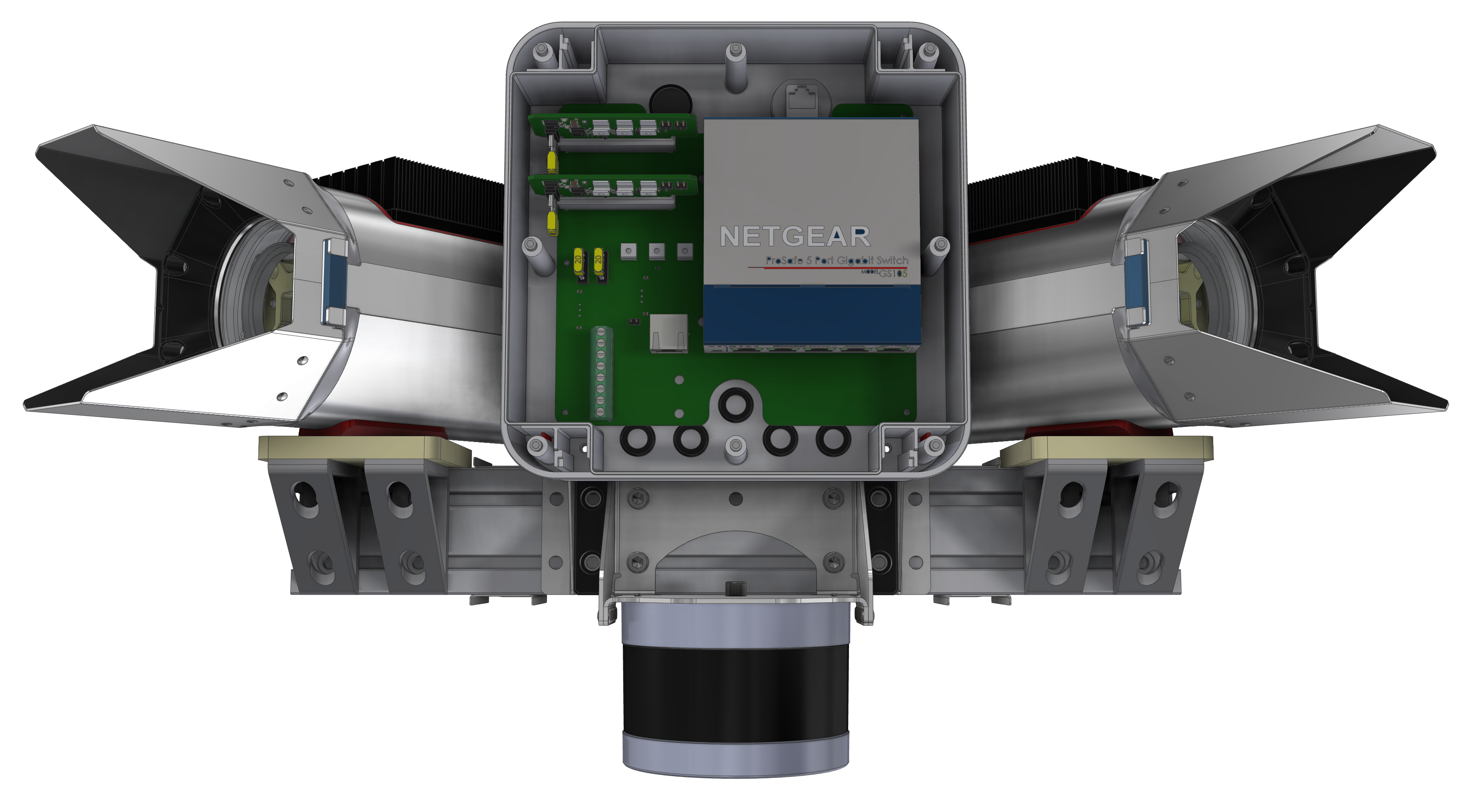}
    }
  \end{center}
  \caption{Front View of Perception Array}
  \label{fig:perception_array_front}
\end{figure}

The foreground and background views of the designed perception array are shown in Fig. \ref{fig:perception_array_front}

\begin{figure}[H]
  \begin{center}
    \scalebox{0.1}{
      \includegraphics{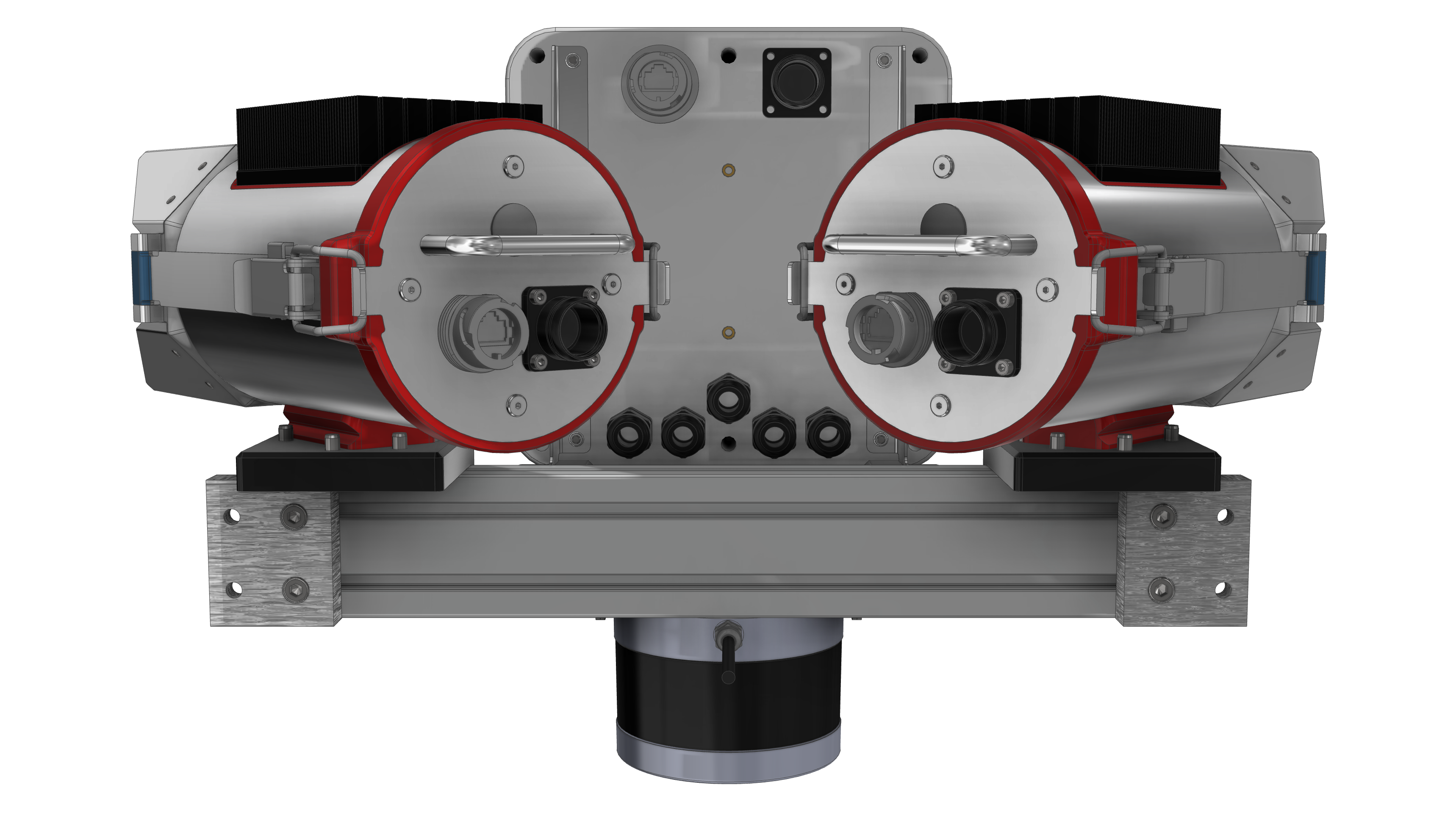}
    }
  \end{center}
  \caption{Rear View of Perception Array}
  \label{fig:perception_array_back}
\end{figure}

and Fig. \ref{fig:perception_array_back}, respectively.

\begin{figure}[H]
  \begin{center}
    \scalebox{0.2}{
      \includegraphics{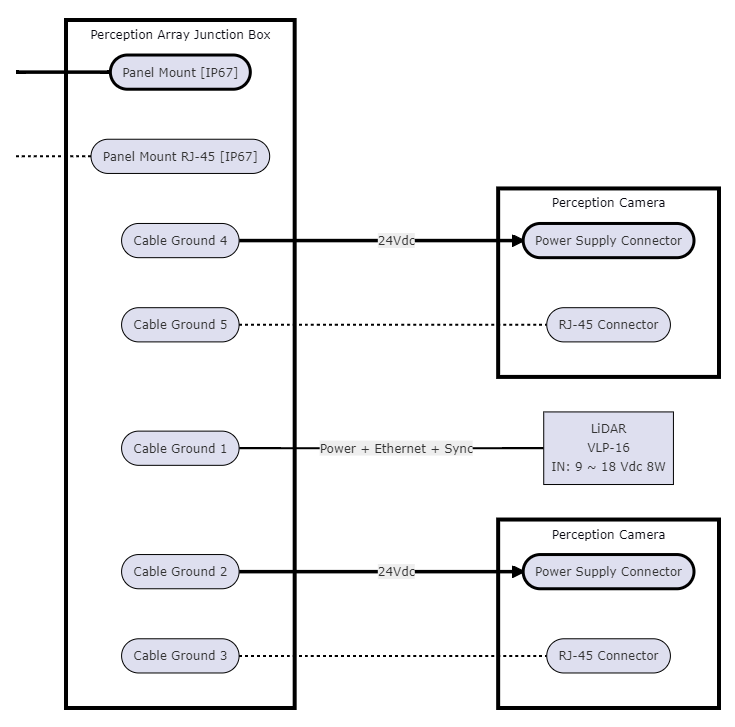}
    }
  \end{center}
  \caption{Diagram of Perception Array}
  \label{fig:perception_aray_diagram}
\end{figure}

The cameras are mounted on the left and right sides, and the LiDAR is mounted upside down at the bottom.
The box in the center contains the power supply function and the switching hub.
This configuration is shown in the Fig. \ref{fig:perception_aray_diagram}.

\subsection{The Perception Camera}

\begin{figure}[htbp]
  \begin{center}
    \scalebox{0.2}{
      \includegraphics{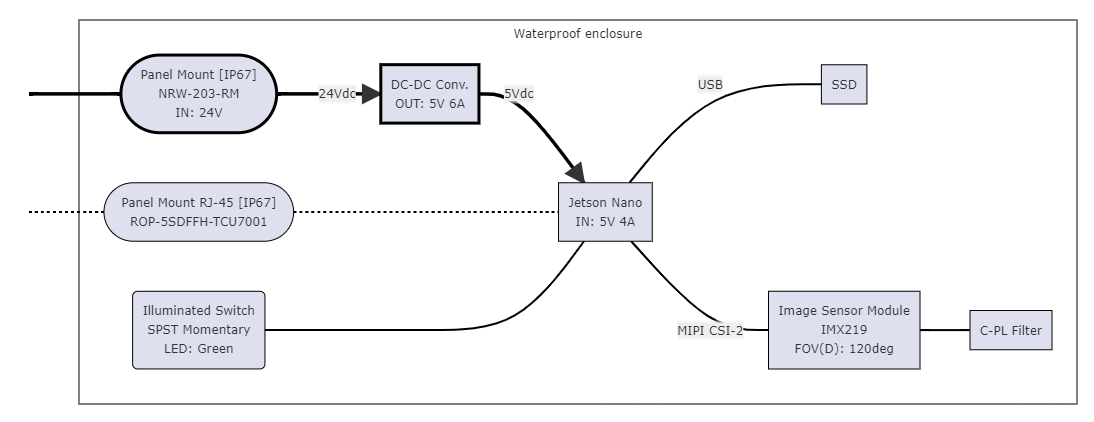}
    }
  \end{center}
  \caption{Diagram of the Perception Camera}
  \label{fig:perception_camera_diagram}
\end{figure}

The camera, which is part of the Perception Array, is designed to perform edge-processing image recognition and is equipped with a Jetson Nano from Nvidia as the computing system.
The camera is designed to be waterproof and heat-dissipating for use in various weather conditions.
The system diagram is shown in Fig. \ref{fig:perception_camera_diagram}.

\begin{figure}[htbp]
  \begin{center}
    \scalebox{0.2}{
      \includegraphics{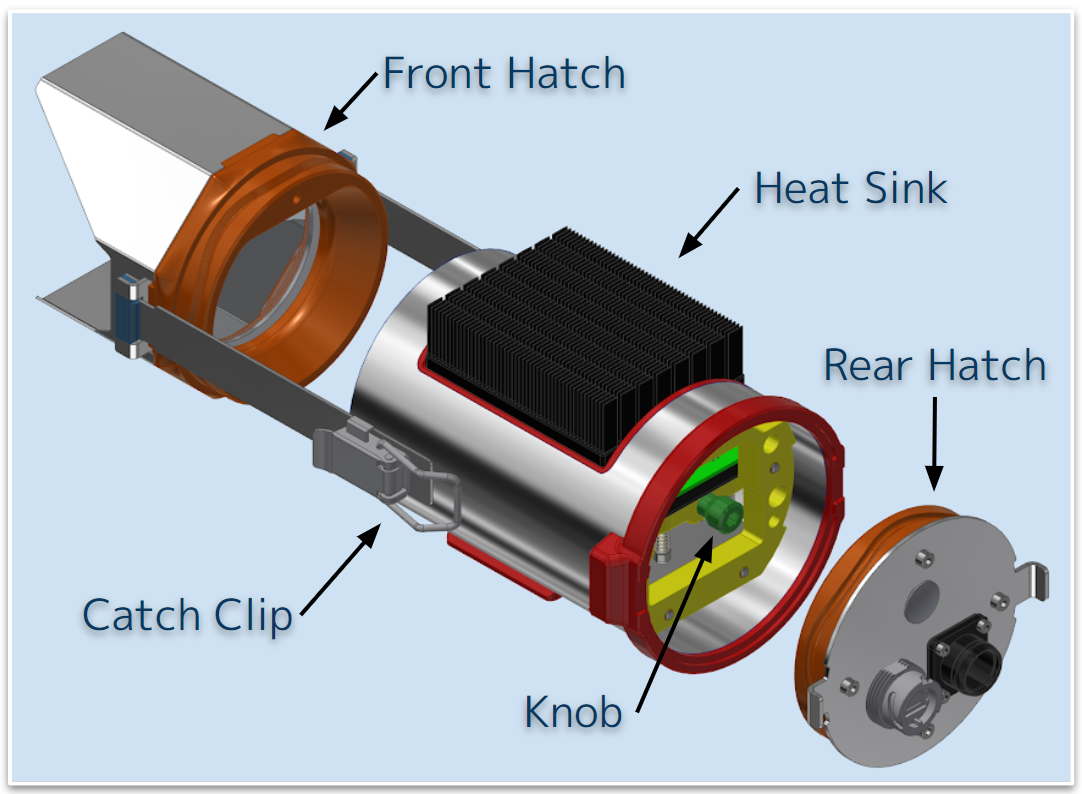}
    }
  \end{center}
  \caption{Disassembled diagram of the Perception Camera}
  \label{fig:camera_disassembly_1}
\end{figure}

The front and rear hatches can be opened and closed without tools. Fig. \ref{fig:camera_disassembly_1} shows these hatches opened.

\subsection{MINI-V in the COVID-19}
MINI-V(minitua vessel) was created in order to test the software easily in the Covid-19. Over the past several years, 
we couldn't conduct the experiment on the ocean or lake because of the COVID-19.
We were prohibited to meet and create the parts of WAM-V.
In addition, the law about vessels is very strict.
So, we can't float the boat easily. The WAM-V is so big and it is hard work and costs too much to carry WAM-V to the lake. 
Then, we need a sustainable system to develop the automotive vessel.
As mentioned above, the simulator is used for developing navigation systems, and it doesn't need to use WAM-V.
The perception array was created to get the sensor data for software tests. They made it easier for us to develop software without ships.
However, the software and hardware integration is the most important to conduct tasks. Then, MINI-V was created 
to make it easier to do the test and the integration.

The concepts of MINI-V are follows:
\begin{enumerate}
  \item easy assembly, transport, and experiment,
  \item open source software and hardware,
  \item high compatibility between WAM-V and MINI-V.
\end{enumerate}

\begin{figure}[H]
    \begin{center}
    \scalebox{0.22}{
        \includegraphics{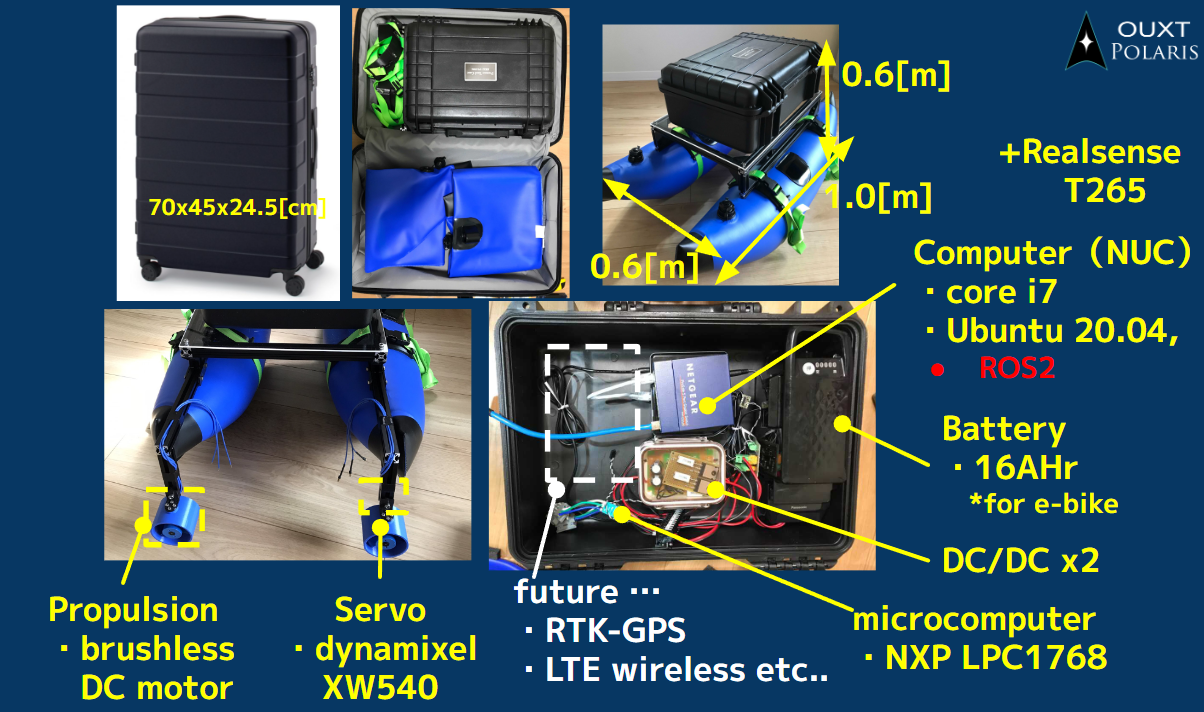}
    }
    \end{center}
    \caption{Hardware Components of MINI-V}
    \label{fig:mini_v_component}
\end{figure}

MINI-V is created to be easy to carry, and we can carry them by suitcase like Fig. \ref{fig:mini_v_component}. It is so small that we can
float and test on the buthtab.
It is also assembled simply. We develop this vessel on open source. So, other people can play or test their software with MINI-V. Finally,
we expect the high compatibility between WAM-V and MINI-V, and it will make it easy to migrate developed software on MINI-V toWAM-V. However, 
MINI-V have not had complete compatibility yet.
We have future tasks to create a little bigger vessel to have compatible hardware and software such as batteries and sensors, and so on.
  
\begin{figure}[H]
    \begin{center}
    \scalebox{0.3}{
        \includegraphics{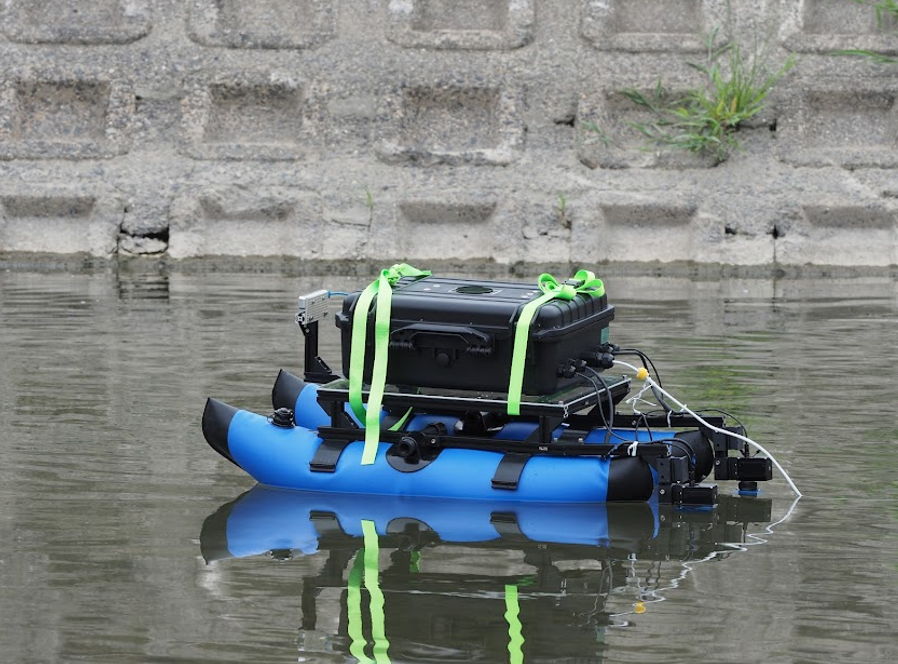}
    }
    \end{center}
    \caption{Experiment on Ai River}
    \label{fig:mini_v_experiment}
\end{figure}

\subsection{the prototype of the multicopter}
In the RobotX 2022, the tasks about multicopter are add on the competition. The drone needs to be automated and pick up the task objects.
We created the quadcopter below in Fig.\ref{fig:drone}.
The flight controller, Pixacer Pro\cite{pixracer}, is introduced to controll the drone. 
In order to get the drone position, it has GNSS sensor on the plane.
we conduct the test of estimating the self-position of the drone. As you can see Fig.\ref{fig:drone_posi} the drone can estimate
the self-position With an error of $\pm 0.5 \mathrm{m}$ approxmately.
At the moment, the drone is manually controled by a pilot but not automated.
For automation, we need to add computers such as raspberryPi to send the operation into flight controller. 
Moreover, the sensor such as cameras to recognize the task objects are needed.
There are many considerations such as the roter-size, motor power, the size of body, and so on.
We are going to develop the automated navigation system for drone in next years.

\begin{figure}[H]
  \begin{center}
    \scalebox{0.3}{
      \includegraphics{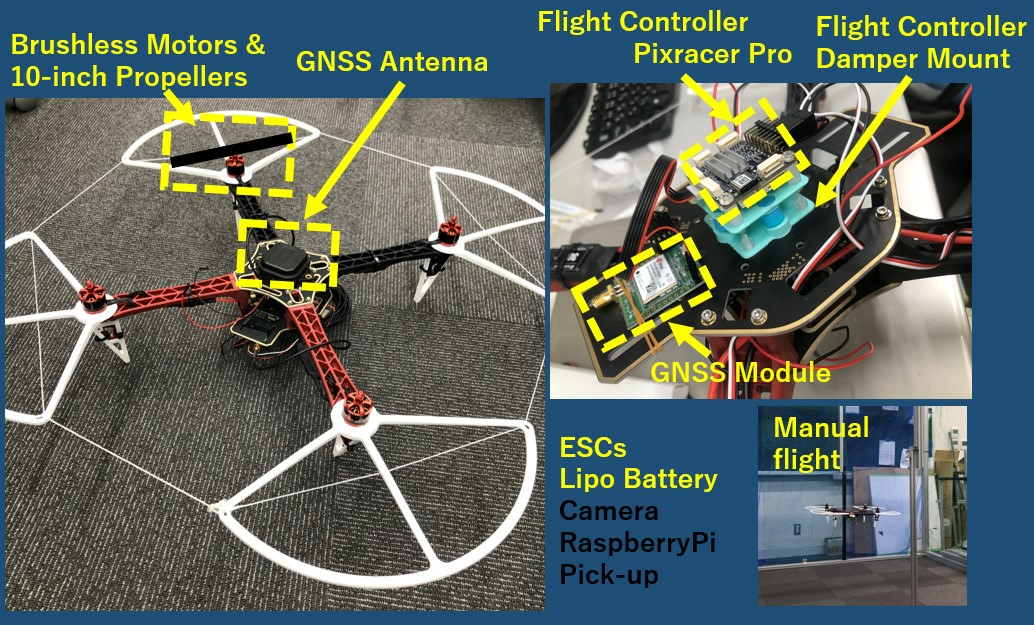}
    }
  \end{center}
  \caption{Componets of Drone}
  \label{fig:drone}
\end{figure}

\begin{figure}[H]
  \begin{center}
    \scalebox{0.6}{
      \includegraphics{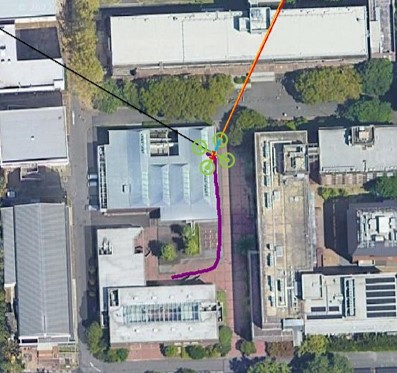}
    }
  \end{center}
  \caption{Estimated Drone Position and Track}
  \label{fig:drone_posi}
\end{figure}

\subsection{Emergency Stop System}

There are four emergency stop switches on the outside of the hull, two of the normal switch type and one wireless one.
When any one of these switches is turned on, the propulsion system is turned off and the ship comes to a safe stop.

\begin{figure}[H]
    \begin{center}
        \scalebox{0.2}{
            \includegraphics{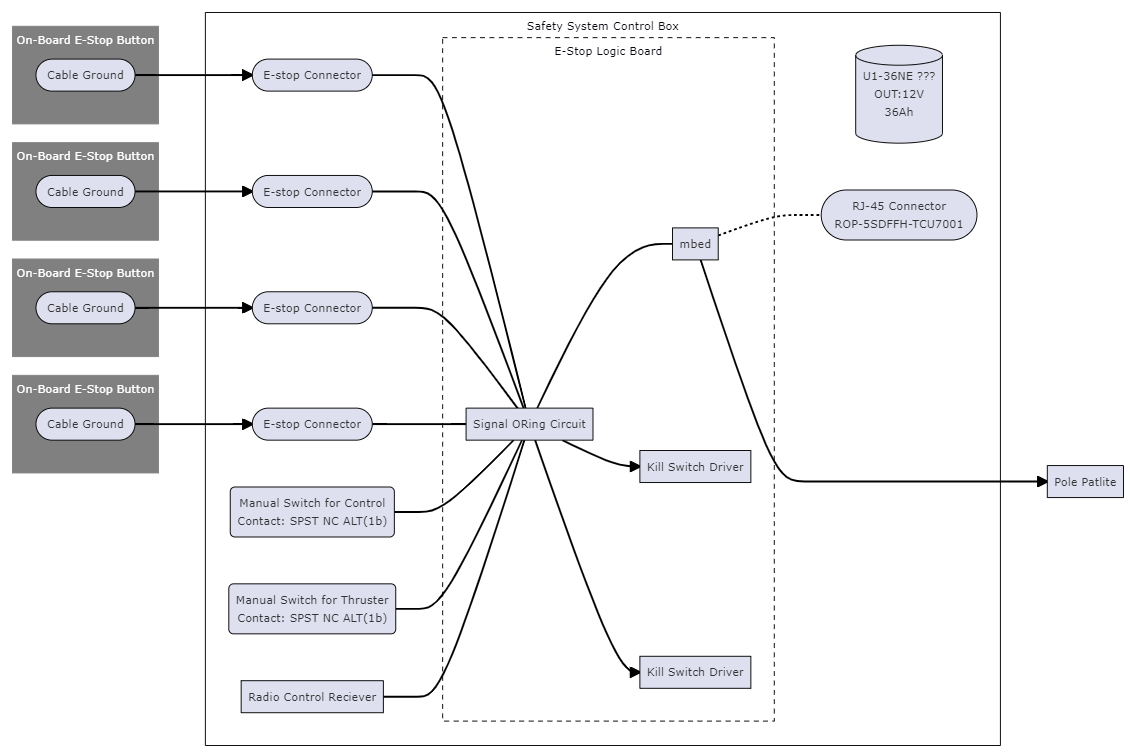}
        }
    \end{center}
    \caption{Emergency Stop System Diagram}
    \label{fig:emergency_stop_system_diagram}
\end{figure}

\begin{figure}[H]
    \begin{center}
        \scalebox{0.25}{
            \includegraphics{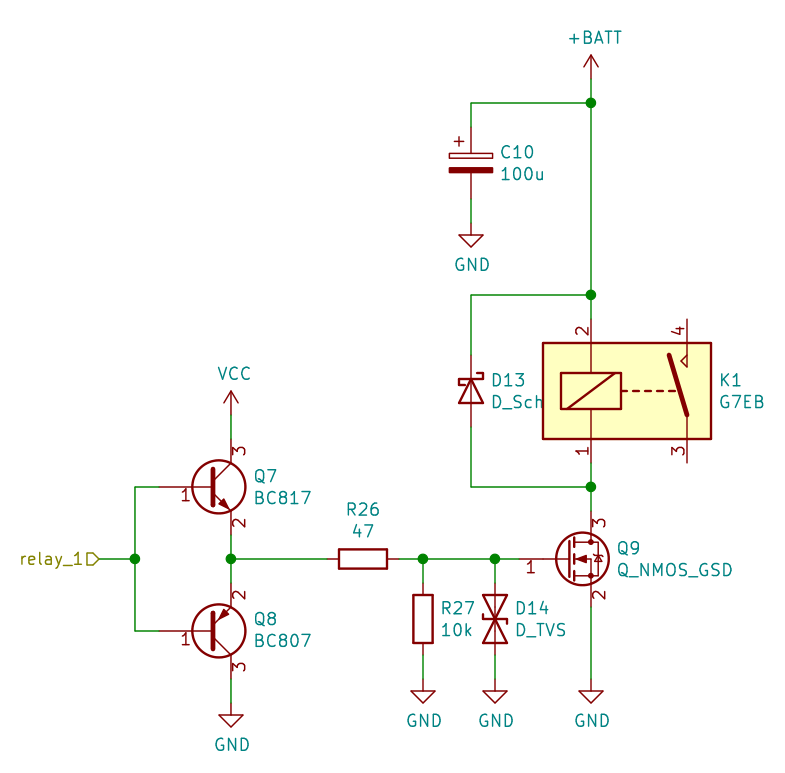}
        }
    \end{center}
    \caption{Kill Switch Relay Driver Circuit}
    \label{fig:kill_switch_relay_driver_circuit}
\end{figure}

\begin{figure}[H]
    \begin{center}
        \scalebox{0.2}{
            \includegraphics{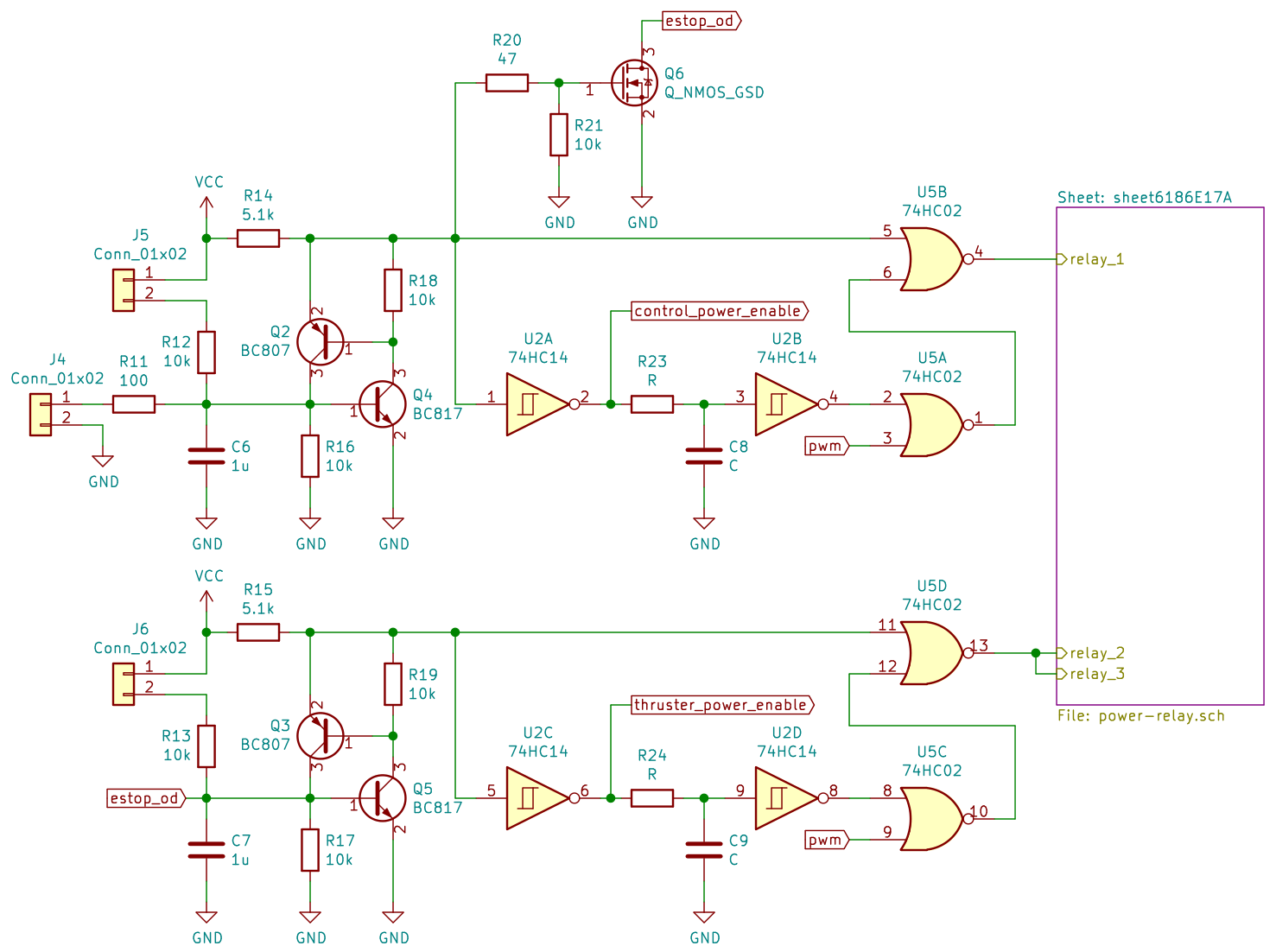}
        }
    \end{center}
    \caption{Kill Switch Signal Circuit}
    \label{fig:kill_switch_signal_circuit}
\end{figure}

In a normal design, a relay used in such a case would drain about 10 W from its coil alone, which is a very high cost for OUXT-Polaris, which does not have sufficient battery capacity.
To avoid this, we designed a circuit that would function reliably as a kill switch when necessary while reducing the power consumption of the relay coil.
Specifically, we selected a relay that applies voltage only at the moment it is turned on and consumes little power at other times, and designed a circuit to accommodate this.
This design achieved a significant reduction in power consumption and greatly extended the cruising time.
In addition, a latch circuit was introduced so that only momentary switches can be operated.
This eliminates the possibility that the emergency stop switch, once pressed,
would automatically deactivate the emergency stop status in the unlikely event that the contacts are unexpectedly removed.

\section{Software Developments}
\subsection{ROS2-based Autonomous Navigation Stack}
In “Maritime RobotX Challenge 2018”, we used Robot Operating System 1(ROS1) for developing software.
However, the development of ROS1 was finished with python2 end of life.
Therefore, we adopted the next generation of ROS called “ROS2”. \cite{ROS2_paper}
As shown in Fig. 1, our ROS2-based simulation and software system was already developed.
Our software contributions are listed as follows.

\begin{itemize}
  \item {\it Software System }:
    We rebuilt the software system from ROS1 to ROS2
  \item {\it Behavior Tree}:
  We adopted the behavior tree library in ROS2 and built our original behavior tree.  
  \item {\it Camera LiDAR object detection}:
    We are developing a lidar-camera fusion object detection system for this project.
  \item {\it Simulation Tool Development }:
    We developed LiDAR simulation by using intel ray-tracing OSS "Embree".
  \item {\it Infrastructure}
    We developed some automation tools for develop quickly.
\end{itemize}
We published all codes in GitHub to give feedback knowledge to the ROS community and 
Open-Source all our resources not only software \cite{documentation_software}
but also CAD models, and circuit data. \cite{documentation_hardware}

\subsection{Software System Architecture}
\begin{figure}[H]
    \begin{center}
        \scalebox{0.12}{
            \includegraphics{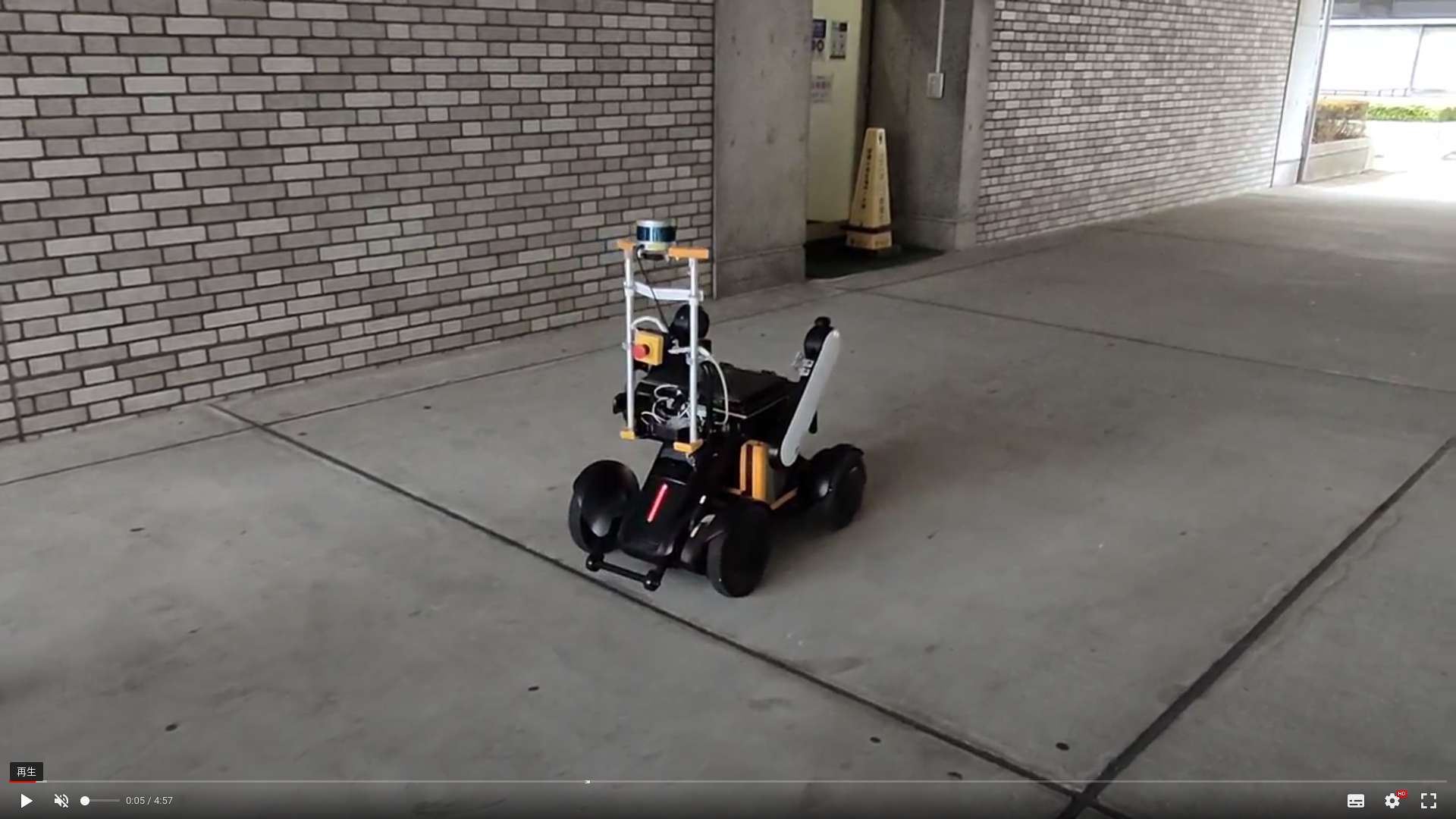}
        }
    \end{center}
    \caption{Autonomous Mobility with Our Navigation Stack}
    \label{fig:whill}
\end{figure}

\begin{figure}[H]
    \begin{center}
        \scalebox{0.12}{
            \includegraphics{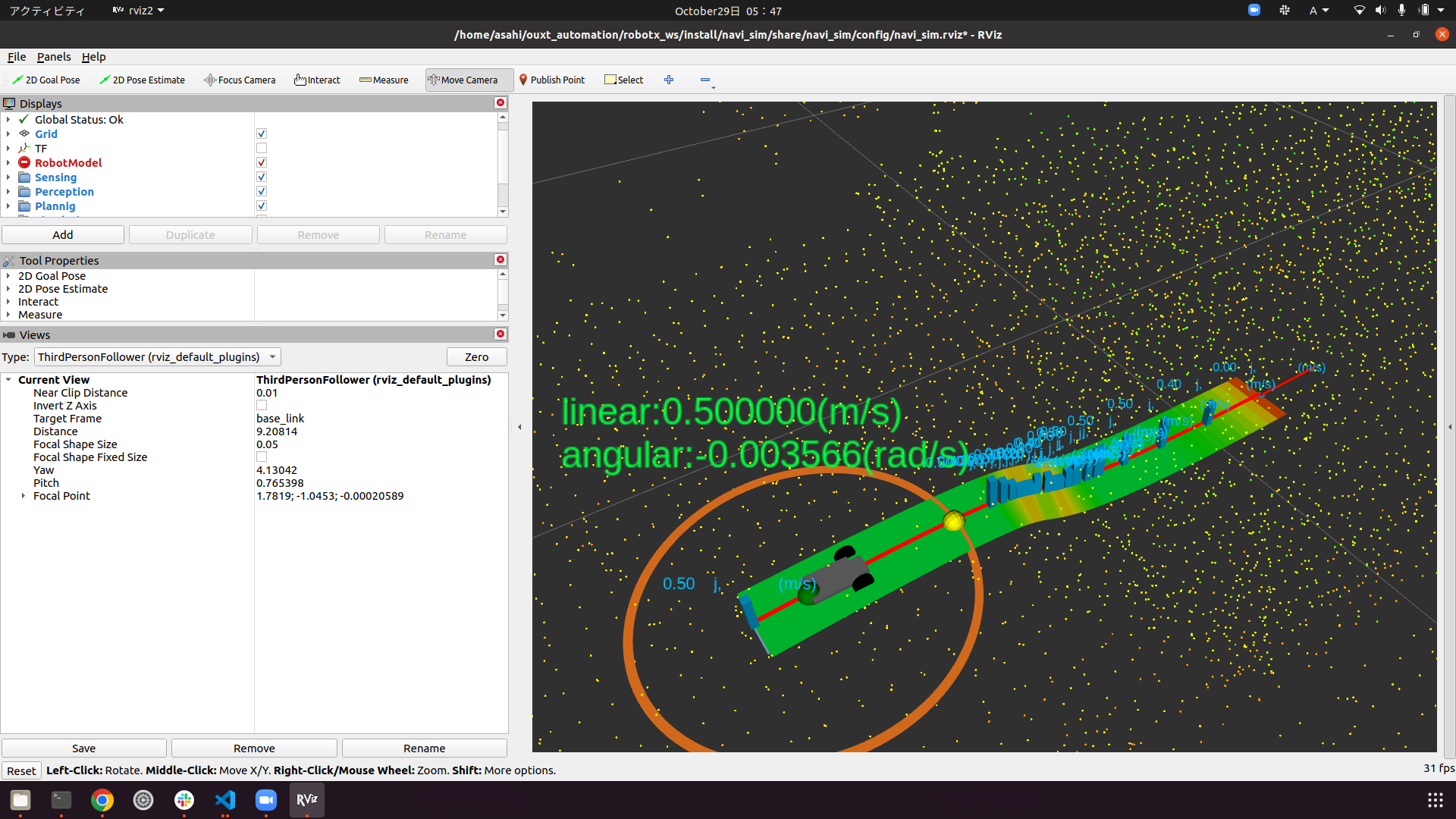}
        }
    \end{center}
    \caption{Autonomous Mobility with Our Navigation Stack (rviz)}
    \label{fig:whill_rviz}
\end{figure}

Our navigation stack is based on ROS2, but we do not use the navigation2 library. We develop our original software.
Our software is highly modularized, so some of our team members use our stacks in other autonomous mobility competitions. 
(Fig.\ref{fig:whill},Fig. \ref{fig:whill_rviz})

All of our software is managed by ansible and can be deployed on real machines without the need for human intervention.
It is open source, including the setup tools \cite{ouxt_automation} and it's documentation \cite{documentation_software},
so anyone in the world can try out the OUXT-Polaris navigation stack.

\subsection{Behavior Tree}
\begin{figure}[H]
  \begin{center}
    \scalebox{0.24}{
      \includegraphics{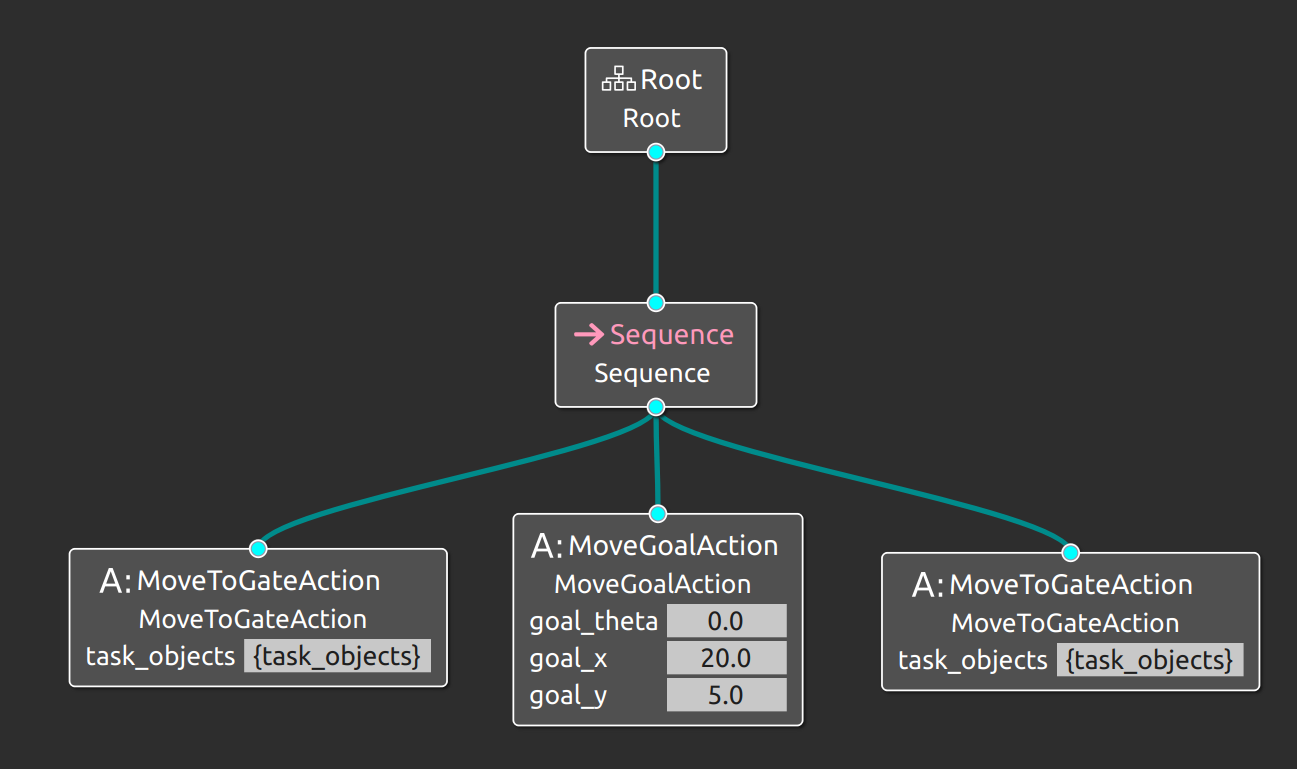}
    }
  \end{center}
  \caption{Groot Example}
  \label{fig:Groot}
\end{figure}

Using behavior tree can build WAM-V behaviors like a tree.  
We use Groot \ref{fig:Groot}, GUI tools for designing behavior tree for smooth development.

For example, in the WAM-V Dynamic Qualifying Task, 
we defined a node that navigation and searches channel markers.
In searching channel markers behavior, first, we got the information from the result of lidar-camera fusion object detection system. 
Information we can get from the system is objects label, probabilities, and bounding boxes.
And we convert the information we can get from the system into information containing channel markers information 
we can use in robotx challenge contains such as labels and coordinates in the same coordinate system of WAM-V etc.
After search channel markers, WAM-V starts navigation.
We need to act the Node two times in the task. So we defined two same nodes and connected them by Sequence node. 
The sequence node is defined in the behavior tree.
We made other Nodes, for example, move forward node, stop node and rotate around the buoy node.
These nodes can realize many behavior patterns required in tasks.

\begin{figure}[H]
    \begin{center}
      \scalebox{0.12}{
        \includegraphics{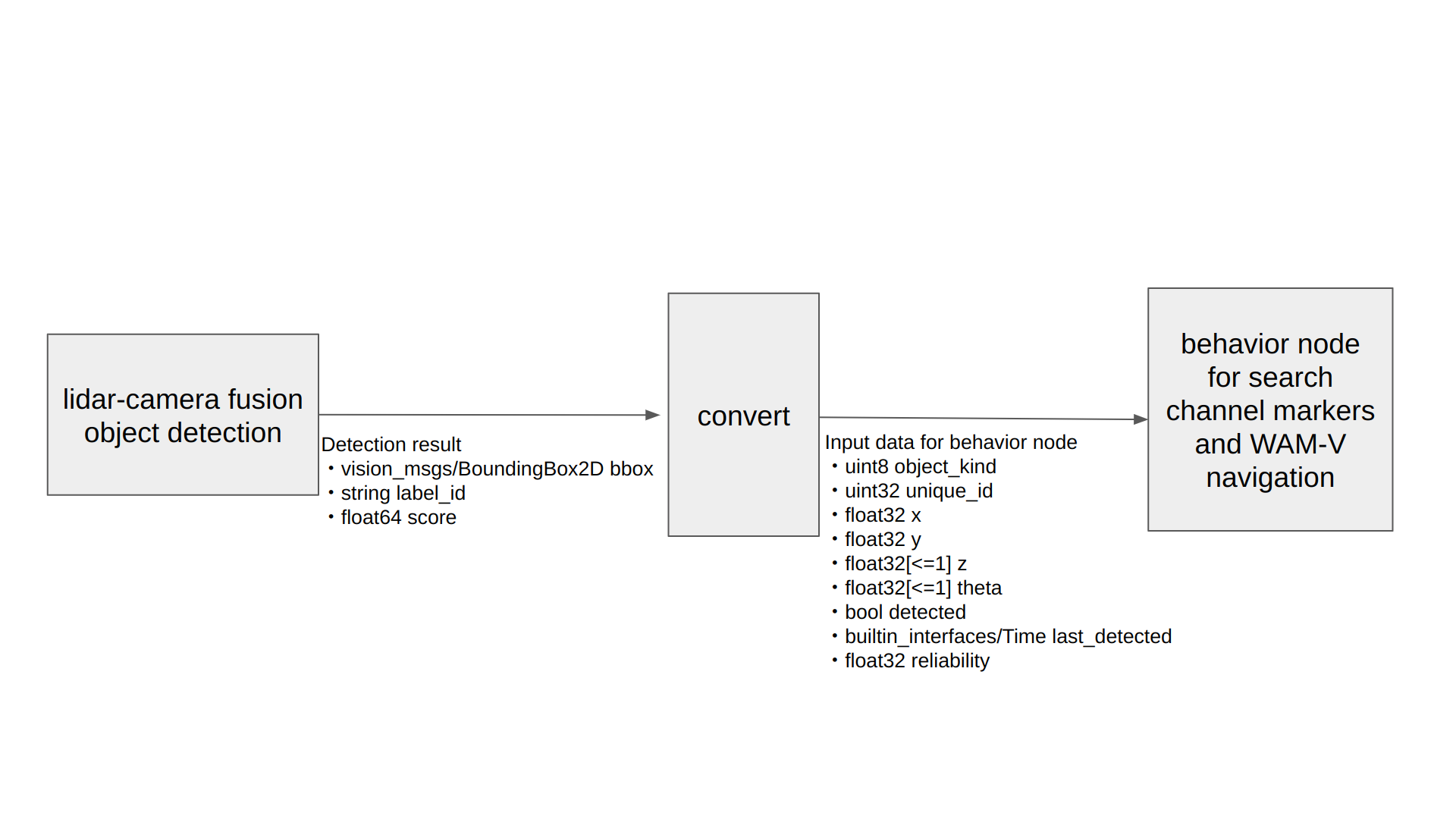}
      }
    \end{center}
    \caption{Pipeline between Perception to Behavior}
    \label{fig:behavior_flow}
  \end{figure}

\subsection{Planner components}
\begin{figure}[H]
  \begin{center}
    \scalebox{0.5}{
      \includegraphics{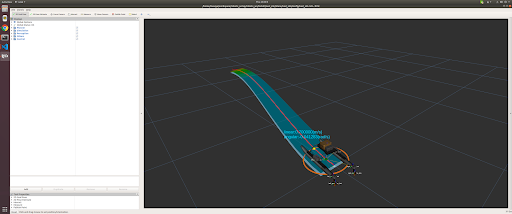}
    }
  \end{center}
  \caption{Part of Planning Results}
  \label{fig:the part of planning results}
\end{figure}

We created hermite planner as the path planning module for WAM-V.
Once the goal direction information from the behavior layer is obtained, the path shape is created using the Hermite curve in combination with the current position.
The Hermite curve is a kind of parametric curve with continuous curvature, and its shape is determined by specifying a vector between two endpoints.
The reason for adopting Hermite curves is that Catmull-Rom Spline, etc. can be created by connecting Hermite curves, so the creation function can be used repeatedly.
And since WAM-V itself can turn on the spot, WAM-V can follow the path such that angular velocity is large,even if the Hermite planner is created.
\indent The components of hermite planner are explained sequentially.
The local waypoint server calculates the distance between the destination and obstacles in the Fresnet coordinate system.
If a contact is detected on the route, a route is created to reach the destination with a slight shift in the X and Y directions in the world coordinate system.
For contact judgment in the Fresnet coordinate system, the nearest neighbor points between the Hermite curve and the 2D LaserScan obtained from the obstacle are calculated using the Newton method.
From among them, the collision between WAM-V and the obstacle objects is judged to have occurred if Satisfying following equations.
\begin{equation}
  0<t<1
\end{equation}
\begin{equation}
  f(t)<width+margin
  \end{equation}
\begin{equation}
  f(t)=at^3+bt^2+c^2+d
  \end{equation}
When the Hermite curve is determined, the calculation of the velocity constraint is performed using several velocity modules.
The first is stop planner, which creates a speed commitment to stop at the destination end of the route.
This speed module creates a speed commitment to slow down at a constant acceleration before heading to the destination.
Second, there is an obstacle planner to perform a stop before obstacle objects.
This constraint to stop before the obstacle uses the same method as stop planner.
The third is a curve planner to prevent the angular velocity from exceeding a certain value.
It creates a velocity constraint so that the angular velocity between points on the created Hermite curve does not become too large.
Each of these three creates velocity constraints independently, integrates the information from each of them.
And, the graph adjusts the velocity constraints, and searches the velocity.
Creating a velocity plan on the route by deleting edges with too much acceleration as edges with increasing or decreasing velocity between them as nodes on each curve.
Then, based on the created route and the ship's own self-position, the pure pursuit planner is used to follow the route.
A contact point between the circle centered on the WAM-V's self-position and the path is created, and the angular velocity and speed up to that contact point is calculated.
In addition, a straight line is created on the extension of the endpoint to approximate the endpoint so that the endpoint does not run out of control.
Fig.\ref{fig:the part of planning results} is shown as an the part of planning results by the above path planning module.

\subsection{Camera LiDAR fusion}
\begin{figure}[H]
    \begin{center}
      \scalebox{0.13}{
      \includegraphics{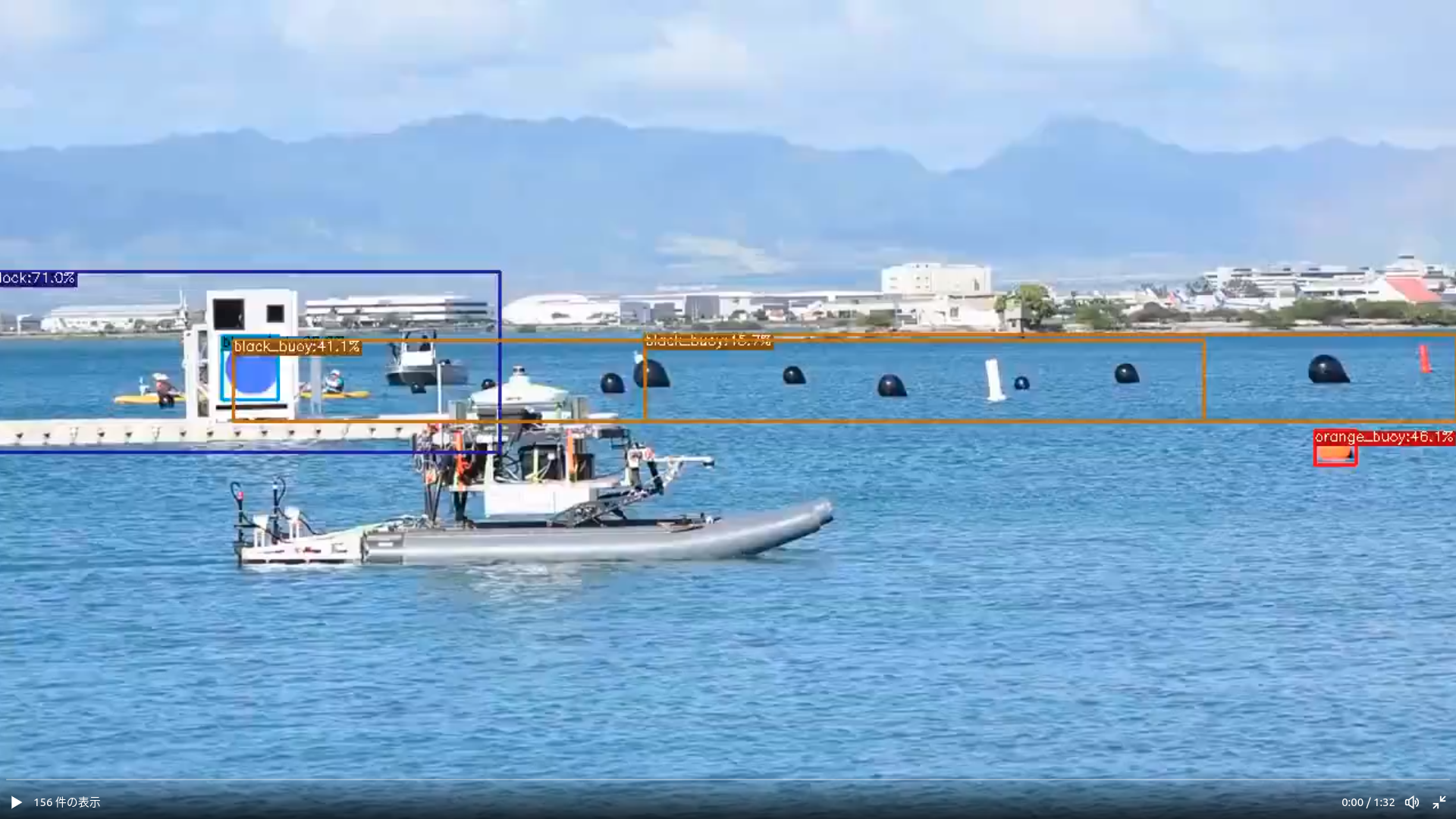}
    }
  \end{center}
  \caption{Inference}
  \label{fig:inference}
\end{figure}

We used  YOLOX\cite{YOLOX} for object Detection of task object information such as buoys and docks.
We created annotation data based on images and videos obtained from past RobotX Challenges.
We open our annotation data for speeding up future development for Maritime RobotX. \cite{dataset_annotations}
The result of training and inference is shown in Fig.\ref{fig:inference}.

The image is part of a video recorded during our challenging navigation in the 2018 RobotX Challenge.\cite{RobotX2018_video}
This model is based on the YOLOX-S network and converted into a tensorrt model.
So, we can run an object detection model of more than 10 Hz in Jetson nano.

\indent The point cloud from LiDAR preprocesses before fusion the camera image and the LiDAR point cloud.
First, The point cloud from the LiDAR is filtered to remove outliers.
Next, the filtered point cloud is downsampled to a 2D Laser Scan.
Then, the object area is extracted from the 2D LaserScan. \cite{scan_segmentation}
The extraction algorithm determines the object area information based on whether the distance between 
the neighboring LaserScan is less than the threshold value determined adaptively.
The clustered point cloud is projected on the camera images.
And we will match the bounding 2D IOU and the object label on the images by Hungarian method.\cite{hungarian}

\begin{figure}[H]
    \begin{center}
      \scalebox{0.2}{
      \includegraphics{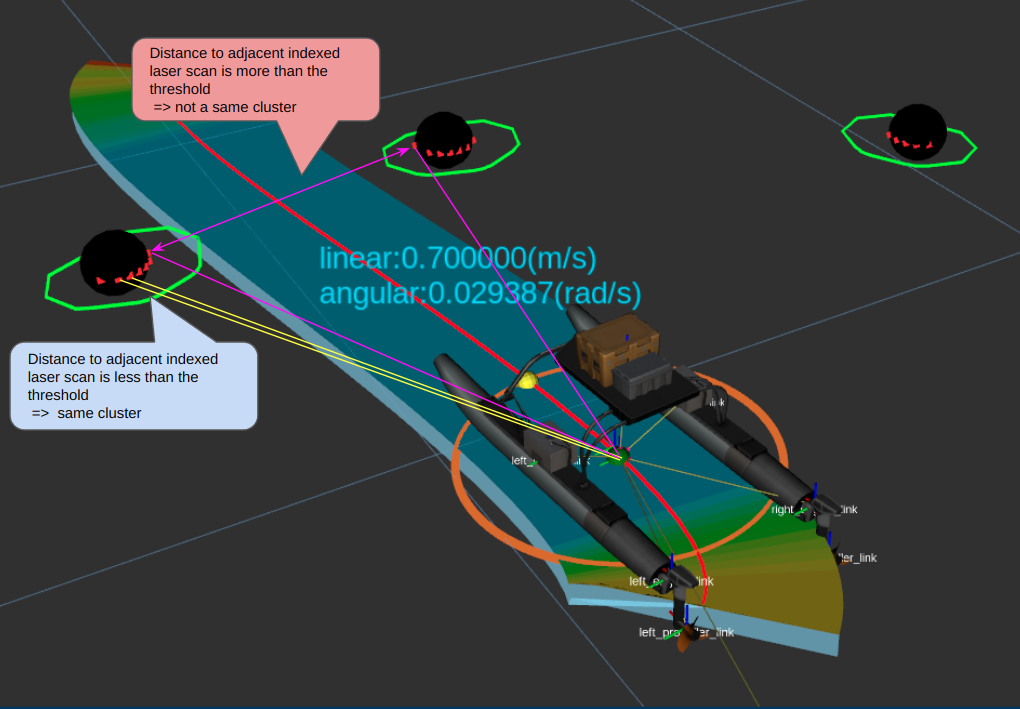}
    }
  \end{center}
  \caption{Algorighum of scan segmentation}
  \label{fig:scan_segmentation}
\end{figure}

\subsection{Simulation}

WAM-V has a car-sized hull, and the setup alone consumes an entire day to conduct the experiment.
The development of a simulator is essential to speed up the development process.
Therefore, we developed a simulator called navi\_sim.
The reason we did not use vrx is that vrx is not yet compatible with ROS2.
navi\_sim utilizes Embree \cite{embree}, an open-source ray tracing library developed by Intel, to simulate lidar with fast CPU ray tracing.
It also provides various other functions such as semantic information output, camera view simulation, and simple ship motion model simulation.

\begin{figure}[H]
    \begin{center}
      \scalebox{0.13}{
      \includegraphics{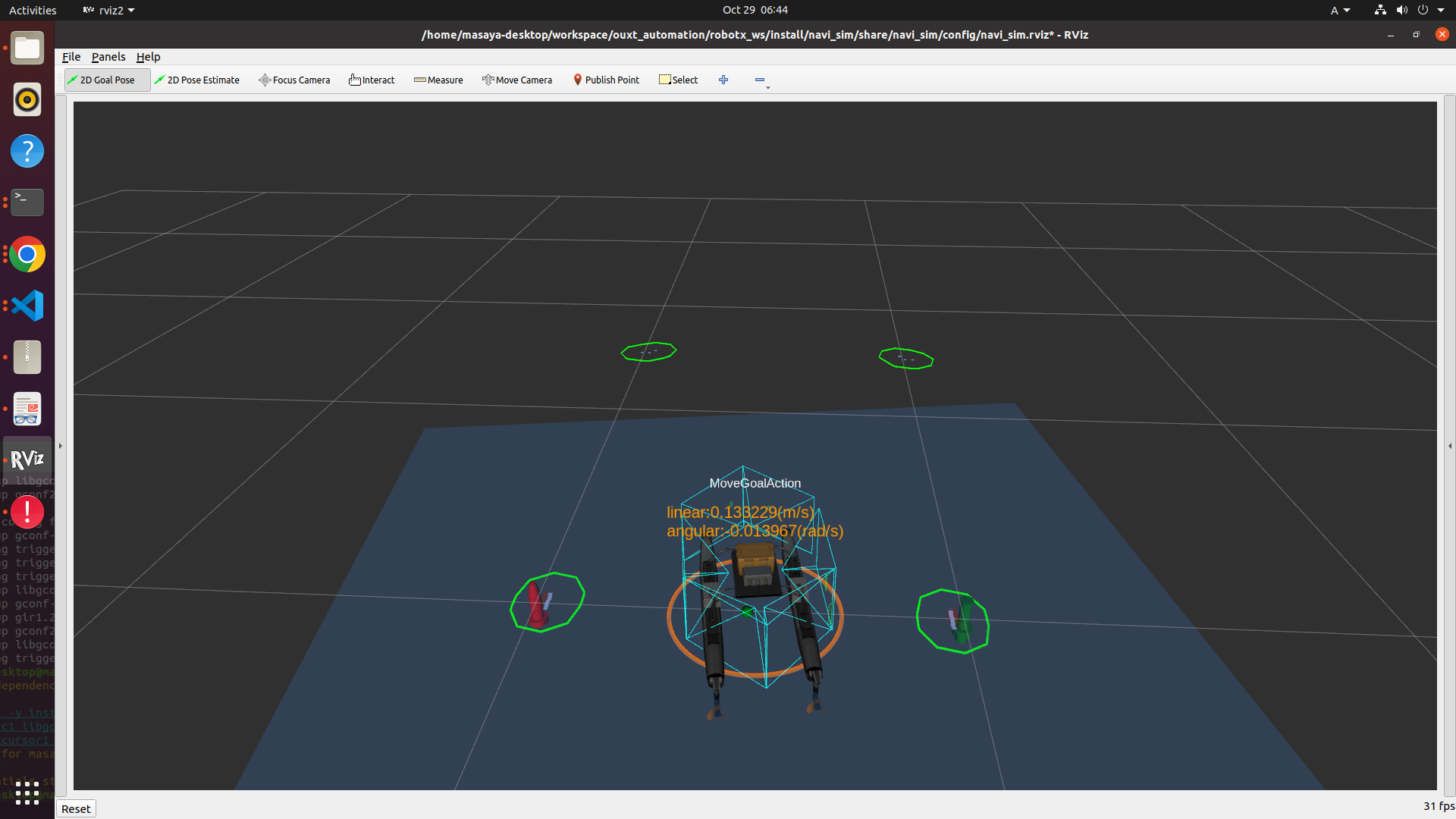}
    }
    \end{center}
    \caption{Running Our Navigation Stack with Navi Sim}
    \label{fig:navi_sim}
\end{figure}

Blue rectangles in Fig. \ref{fig:navi_sim} is camera view simulation, and robot 3D models in \ref{fig:navi_sim} is simulated robot.

\subsection{Infrastructure}
Our system runs in a complex distributed computing environment that includes microcontrollers,
so the configuration file is huge and there are many processes that need to be launched to start the system.
If such a system is deployed manually, various human errors are expected to occur.
In addition, there are 77 ROS2 packages that make up our autonomous navigation system,
and it is almost impossible to manage all of them manually without error.
To solve these problems, we used a configuration management tool called ansible.
ansible can support environment construction procedures on the computer in YAML format and
can switch between development and production environments with a single option.
With the adoption of ansible, setting up our system is just a matter of running a single shell script.

\begin{figure}[H]
    \begin{center}
      \scalebox{0.18}{
      \includegraphics{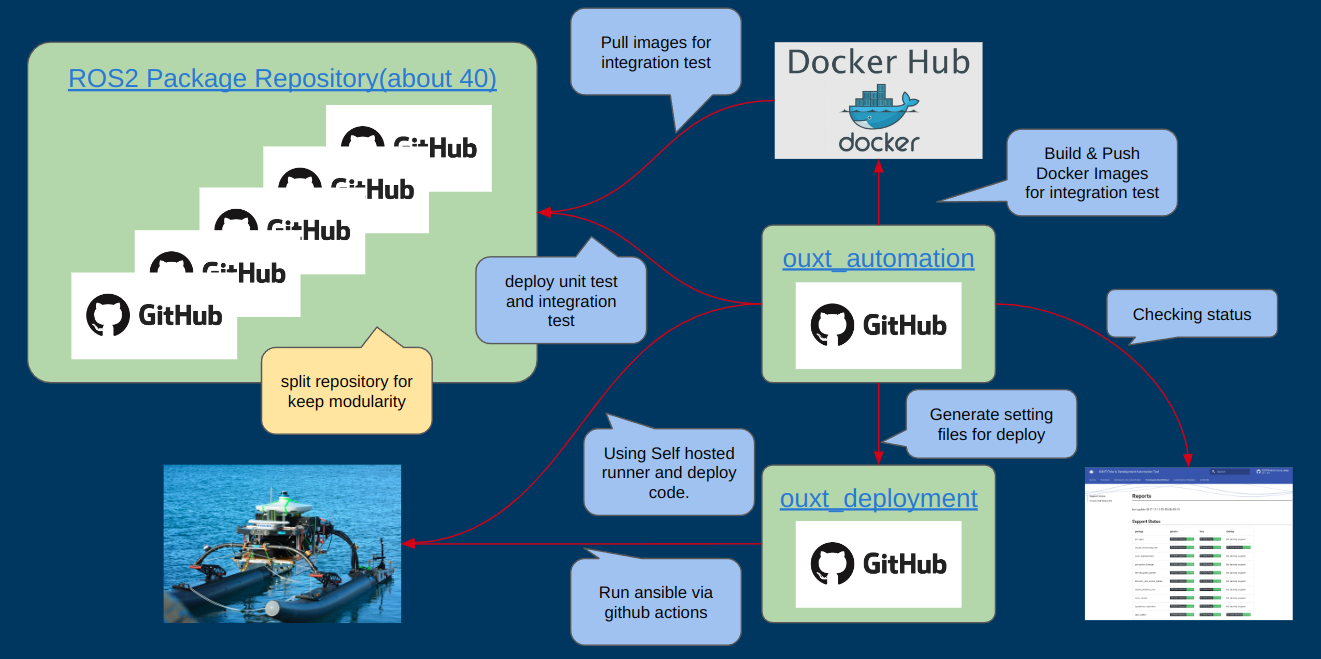}
    }
  \end{center}
  \caption{CI/CD Pipeline}
  \label{fig:ci_cd_pipeline}
\end{figure}

Also, even if the ROS2 package does not have any changes in the code of the package itself,
the software may break if there are changes in the packages on which the package depends.
To detect and fix this quickly, we built a CI system using GitHub Actions.
CI is a technology called continuous integration, which detects commits to Github, etc.,
and automatically performs tests to find defects early.
When a pull request is issued for any of the packages we manage, a build test is automatically performed once a day.
If the build test fails, the pull request cannot be merged.
Failed build tests are notified to the OUXT-Polaris Slack so that members know immediately if a failure has occurred.
Whole architecture of CI/CD pipeline shown in Fig.\ref{fig:ci_cd_pipeline}.
The robot also consists of a single-board computer with an arm CPU as well as a CPU with x86-based architecture,
and a microcontroller with an arm CPU and mbed OS.
The infrastructure supports all of these architectures and 
can test all software from deep-learning recognition systems to low-layer control systems.

\begin{figure}[H]
    \begin{center}
      \scalebox{0.15}{
      \includegraphics{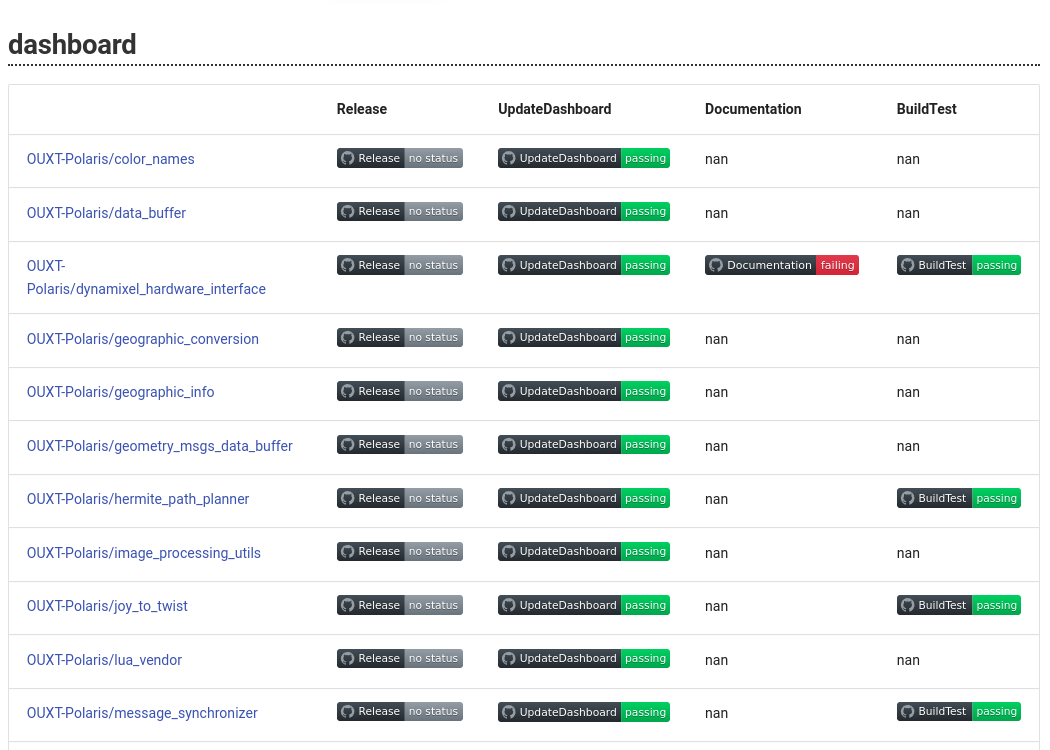}
    }
  \end{center}
  \caption{CI/CD Dashboard}
  \label{fig:ci_cd_dashboard}
\end{figure}

We divide our packages into smaller pieces in order to maintain the high degree of component nature of the packages we develop.
This reduces unnecessary build targets and greatly reduces CI time (tests that originally took 30 minutes or more are now only 3 minutes).
However, since there are more than 40 packages under our control alone,
it is impossible to keep track of their development and testing status without tools.
Therefore, we built a system that calls the API of GitHub and automatically creates a dashboard in Fig.\ref{fig:ci_cd_dashboard}.
All of these deliverables are also open-source software.
We also developed GitHub Actions to configure GitHub Actions to unify CI procedures for multiple repositories and
built a system to synchronize CI procedures at all times via bot accounts. \cite{wam-v-tan_bot}

GitHub Actions is also used when deploying to the actual machine.
GitHub Actions has a function called Self-Hosted Runner, which allows you to remotely execute any command using your own computer.
Using this function, after the CI of the ROS2 package is completed, 
a configuration file with all commit hashes fixed is created and deployed to the actual device
using Github Actions/ansible based on that configuration file.
This allows our team to deploy the verified and up-to-date source code to the machine at the push of a button.

In addition, machine learning is a task that requires a variety of complicated labor and a machine with a high-performance GPU.
However, most tasks are routine and human intervention is not productive.
To give some concrete examples, the following tasks can be automated.

\begin{figure}[H]
    \begin{center}
      \scalebox{0.1}{
      \includegraphics{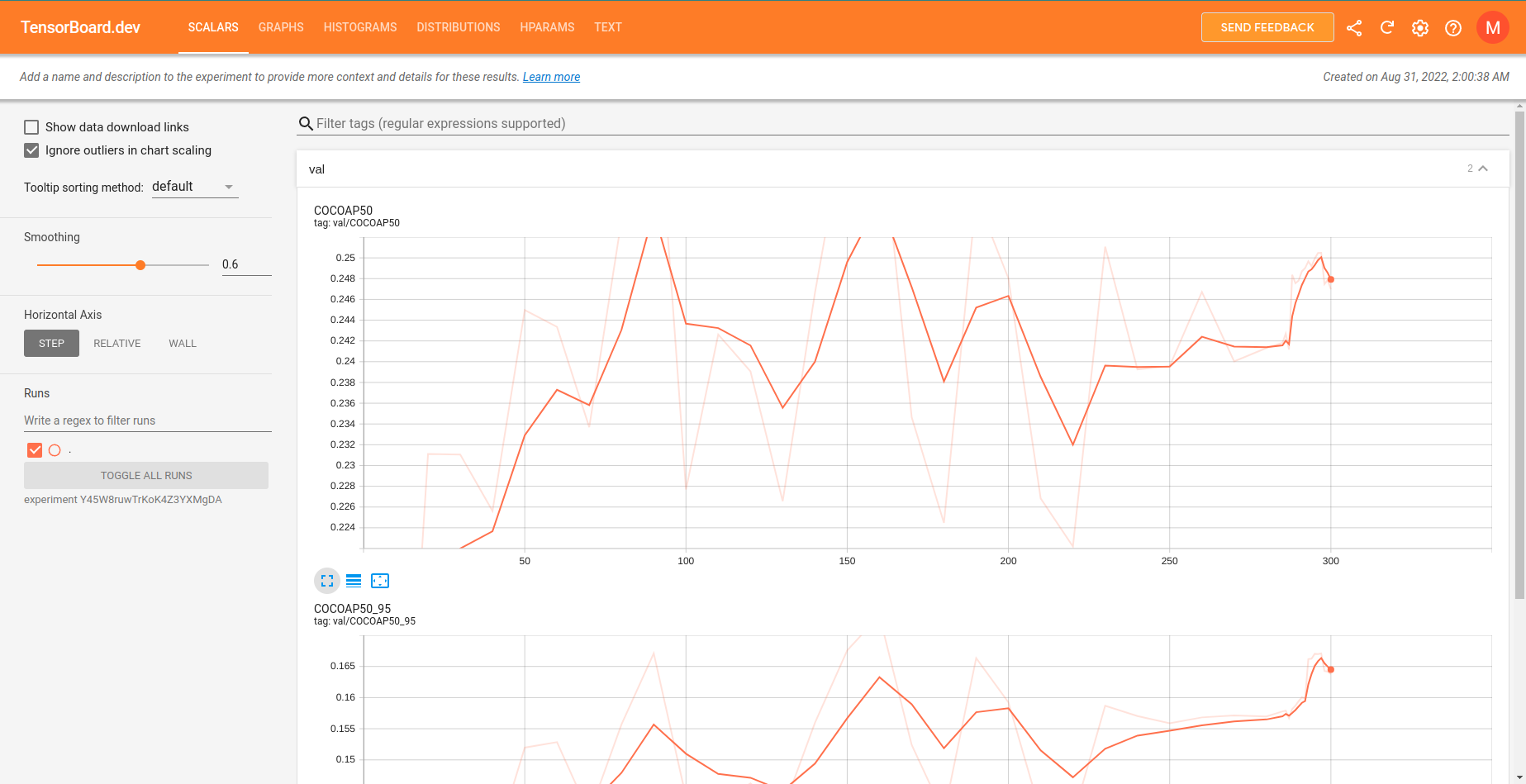}
    }
  \end{center}
  \caption{Training Result in Tensorbord.dev}
  \label{fig:tensorbord}
\end{figure}

\begin{itemize}
    \item {\it Conversion of dataset formats }:
        We use labelme for annotation tools, but YOLOX use coco format for learning.
    \item {\it Performing training }:
        Virtualize the environment with nvidia docker and run scripts for learning on it
    \item {\it Visualization and versioning of training results }:
        We can visualize inference result via Twitter bot(Fig.\cite{wam_v_tan_bot}) and tensorbord.dev(Fig.\ref{fig:tensorbord})
    \item {\it Model Conversion }:
        Need to convert PyTorch training results into a tensorrt model for faster inference.
\end{itemize}

All tasks are executed using the Self-hosted runner in GitHub actions, 
which automatically executes the training when it detects additional events in the machine learning data set.
Training takes about 30 minutes when we train YOLOX-S model.
Future work includes building a mechanism to automatically deploy the obtained learning results to 
the actual machine and Neural Architecture Search using black-box optimization.

\subsection{Communication with Technical Director Server}
We must report the vehicle's status as the Autonomous Maritime System heartbeat to a Technical Director server via a TCP connection. 
Our team usually tests navigation systems using a simulation environment, 
so we should check the heartbeat sentence on the simulation loop. 
To achieve it, we built a simple mock server\cite{robotx_communication}.
It displays received NMEA-like sentences from a simulation via a TCP connection.
We also implemented a ROS2 node that collects information about the aircraft and sends heartbeats.

\subsection{WAM-V Controller}
The WAM-V control system adopted the ros2\_control\cite{ros2_control,wamv_control}.
The ros2\_control is a framework for real-time control of robots using ROS2.
Our WAM-V controllers are listed as follows.
\subsubsection{Equation of Motion}
Our WAM-V equation of motion is calculated as follows.
\begin{equation}
  M \dot\nu = -D(\nu)\nu + \tau
\end{equation}

\begin{equation}
  \nu = 
  \begin{bmatrix}
  u & v & \omega \\
  \end{bmatrix}^T\\
  \label{eq:control_1}
\end{equation}

\begin{equation}
\tau = 
\begin{bmatrix}
f_x & f_y & f_{yaw} \\
\end{bmatrix}^T\\
\end{equation}

\begin{equation}
M = 
\begin{bmatrix}
m+m_x &  0 & 0 \\
0 & m+m_y & 0 \\
0 & 0 & I_z + J_z
\end{bmatrix}\\
\end{equation}

\begin{equation}
D(\nu) = 
\begin{bmatrix}
0 & 0 & -mu \\
0 & 0 & mv \\
mu & -mu & 0
\end{bmatrix}\\
\end{equation}

where $\nu$ is WAM-V velocity in WAM-V coordinate system.
$\tau$ is WAM-V coordinate system input.
$M$ is the mass matrix.
$m$ is the WAM-V mass.
$m_x$ and $m_y$ are the added mass of WAM-V in X and Y directions.
$I_z$ is the WAM-V moment of inertia.
$J_z$ is the WAM-V added moment of inertia.
$D(\nu)$ is the WAM-V resistance coefficient matrix.

\subsubsection{Differential Thruster Model}
\begin{eqnarray}
\tau &=& 
\begin{bmatrix}
f_x \\ 
f_y \\
f_{yaw} \\
\end{bmatrix}\nonumber  \\
&=&
\begin{bmatrix}
\dfrac{1}{2}(F_r + F_l) \\
0 \\
\dfrac{w}{2}(F_r - F_l) 
\end{bmatrix} \\
\nonumber  
\end{eqnarray}

where $Fr$ and $F_l$ are propulsion of right and left side thruster.
$w$ is the distance between right and left thrusters.

\subsubsection{Propeller Model}
\begin{eqnarray}
T &=& \rho n^2 D^4 K_t(J_s)\\
K_t(J_s) &=& k_2J_s^2 + k_1J_s + k_0\\
J_s &=& \dfrac{up}{nD}
\end{eqnarray}

where,
$T$, $\rho$, and $D$ are the propeller thrust, fluid density, and propeller radius.
$k_0$, $k_1$ and $k_2$ are constants values.
$up$ and $n$ are inflow rate and rotation speed.

\section{Firmware}
\begin{figure}[H]
    \begin{center}
      \scalebox{0.23}{
      \includegraphics{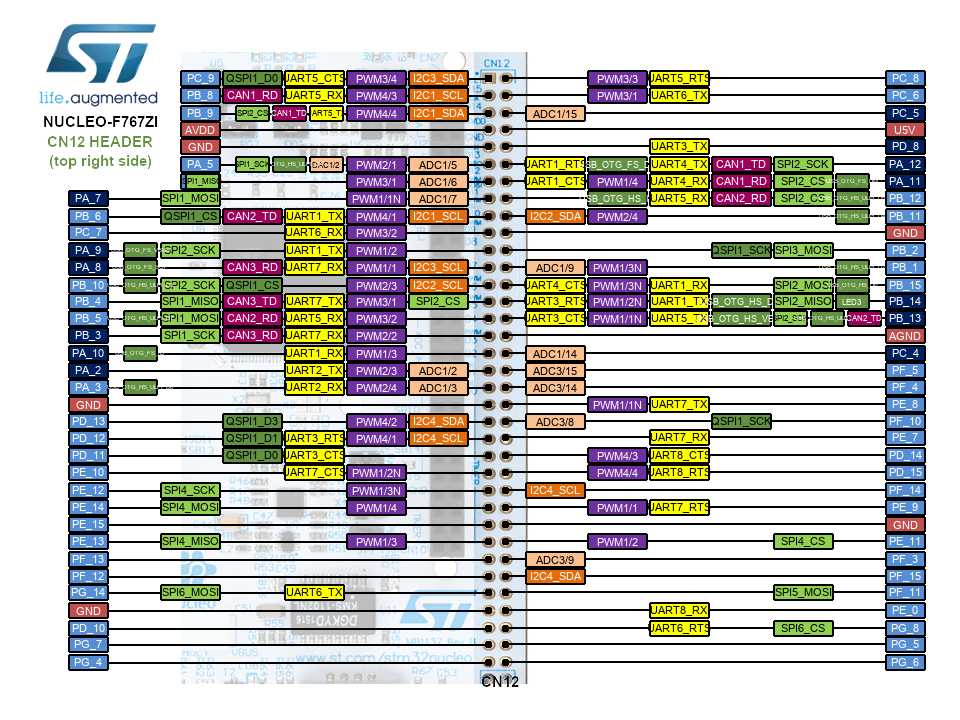}
    }
  \end{center}
  \caption{NUCLEO-F767ZI Microcontroller}
  \label{fig:f767zi}
\end{figure}

We use the same microcontroller for two main purposes.
The reason for using the same microcontroller is to make it easier to have a spare in case of failure during the competition.
We considered several types of microcontrollers and finally adopted NUCLEO-F767ZI (Fig.\ref{fig:f767zi}) from STM Corp.

\begin{table}[H]
    \caption{Board Selection}
    \label{table:board_selection}
    \centering
     \begin{tabular}{clll}
      \hline
      board & ROS2 Integration & Ethernet & Serial \\
      \hline \hline
      NUCLEO-F767ZI & mROS2 & Yes & Yes \\
      Teensy 4.0 & micor-ROS & No & Yes \\
      Arduino Due & micro-ROS & No & Yes \\
      LPC1768 & raw-TCP/IP & Yes & Yes \\
      \hline
     \end{tabular}
\end{table}

The two roles of the microcontroller are to drive the motor and monitor the supply voltage.
In order to meet the specifications for designing a real-time guaranteed control system for the motor drive,
we implemented firmware that connects ROS2 through packet communication with the speed control system and UDP communication,
which were specified in the previous chapter.
For the other role of monitoring the power supply voltage, a real-time guarantee was not necessary.
Instead, it was necessary to be able to easily communicate with ROS2 via Pub/Sub, so mROS2, an embedded communication library compatible with ROS2, 
was adopted to realize communication between the lower layers and higher layers using the ROS2 system.

\section{Conclusion}
In this paper, OUXT-Polaris reported the development of the autonomous navigation system for the 2022 RobotX Challenge.
Based on the results of the 2018 RobotX Challenge, OUXT-Polaris rebuilt the system and
developed improved systems for the Maritime RobotX Challenge 2022.
We succeeded in constructing the highly reusable system by designing systems with high independence as parts,
in addition to high computing capacity and environmental recognition capability.
Moreover, we described the developing method in Covid-19 and the feature components for the next RobotX Challenge.
We hope these significant upgrades will produce positive results in the next competition.


\vfill

\end{document}